\documentclass[10pt]{article}
\usepackage[preprint]{tmlr}  

\usepackage{booktabs}
\usepackage{amsfonts}
\usepackage{amsmath}
\usepackage{amssymb}
\usepackage{nicefrac}
\usepackage{xcolor}
\usepackage{graphicx}
\usepackage{subcaption}
\usepackage{adjustbox}
\usepackage{multirow}
\usepackage{algorithm}
\usepackage{float}
\usepackage{hyperref}
\usepackage{url}
\usepackage{microtype}
\usepackage{placeins}
\usepackage{needspace}

\usepackage{algorithmic}

\hypersetup{colorlinks=true,citecolor=blue!55!black,linkcolor=blue!55!black,urlcolor=blue!55!black}

\newcommand{\best}[1]{\textbf{#1}}
\newcommand{\secondbest}[1]{\underline{#1}}
\newcommand{\ours}{\mbox{PreDiff-LM}}
\newcommand{\mask}{\texttt{[MASK]}}
\newcommand{\E}{\mathbb{E}}
\newcommand{\R}{\mathbb{R}}
\newcommand{\V}{\mathcal{V}}

\title{\ours{}: Pretrained Discrete Masked Diffusion Language Modeling with
Hybrid Attention\thanks{Language-model tools were used for editorial
assistance. The authors reviewed and take responsibility for all claims,
analyses, and text.}}

\author{
  Zhengtao Yao\textsuperscript{1,}\thanks{Equal contribution.} \quad
  Runhao Li\textsuperscript{1,}\footnotemark[2] \quad
  Xupeng Chen\textsuperscript{2} \quad
  Jiayi Cheng\textsuperscript{2} \quad
  Chenqian Le\textsuperscript{2} \\
  Michael Yue\textsuperscript{3} \quad
  Jesson Wang\textsuperscript{1} \quad
  Siheng Wang\textsuperscript{4} \quad
  Guang Yang\textsuperscript{5} \quad
  Haoyan Xu\textsuperscript{1} \\
  Chenhao Wei\textsuperscript{5} \quad
  Zhengqing Yuan\textsuperscript{6} \quad
  Youran Shen\textsuperscript{4} \quad
  Yanfang Ye\textsuperscript{6} \quad
  Junhao Dong\textsuperscript{7}
  \\[6pt]
  \addr
  \textsuperscript{1}University of Southern California \quad
  \textsuperscript{2}New York University \quad
  \textsuperscript{3}Columbia University \\
  \textsuperscript{4}University of California, Berkeley \quad
  \textsuperscript{5}Stevens Institute of Technology \\
  \textsuperscript{6}University of Notre Dame \quad
  \textsuperscript{7}Nanyang Technological University
}

\begin{document}
\maketitle

\begin{abstract}
Discrete masked diffusion language models support bidirectional generation and
infilling, but adapting pretrained autoregressive (AR) transformers requires
reconciling causal pretraining with bidirectional denoising. We study this
problem at the level of attention rather than claiming AR-weight reuse itself
as novel. \ours{} preserves causal attention within the observed prompt while
allowing full bidirectional attention within the masked target. Under a matched
GPT-2 Medium, WikiText-103, 90K-step setup, this hybrid mask improves
unconditional perplexity from $34.1$ to $28.7$ and MAUVE from $0.71$ to $0.78$
over uniform bidirectional attention with the same AR initialization. Attention
adaptation also composes with a DiffuGPT-style objective adaptation, reaching
$26.9$ perplexity. Pretrained initialization reduces the steps required to
reach perplexity below 50 from about 350K to 8K, although a compute-matched
fine-tuned AR model remains stronger at equal scale ($18.9$ versus $28.7$).
Beyond perplexity, \ours{} improves repetition, distributional quality, four
zero-shot downstream tasks, and human preference over prior diffusion
baselines. The results position hybrid attention as a complementary mechanism
for adapting pretrained causal backbones, while making explicit the remaining
quality and inference-efficiency gaps to optimized AR models.
\end{abstract}

\section{Introduction}
\label{sec:intro}

Autoregressive (AR) language models generate high-quality text by predicting one
token at a time~\citep{radford2019language,brown2020language}. Their causal
factorization is effective for left-to-right generation, but does not natively
support arbitrary-order refinement or infilling. Discrete diffusion language
models instead corrupt and denoise whole sequences
~\citep{austin2021structured,sahoo2024simple,lou2023discrete}. This gives them
bidirectional generation capabilities, but many early systems were trained
from scratch and remained substantially behind comparably sized AR models.

Recent work has shown that this gap is not solely a property of the diffusion
objective. DiffuGPT and DiffuLLaMA adapt pretrained AR models through continual
diffusion training~\citep{gong2024scaling}; Dream initializes a discrete
diffusion LM from an AR-pretrained model and introduces context-adaptive noise
rescheduling~\citep{ye2025dream}. These results make AR initialization an
established strategy. The unresolved question we study is narrower: \emph{how
should information flow change when causal weights are reused for masked
denoising?}

The difficulty is an attention-pattern mismatch. AR weights are learned under
strictly causal attention, whereas masked target tokens benefit from both left
and right context. Replacing the causal mask everywhere with a bidirectional
mask changes the representations seen by every layer, including the observed
conditioning prompt. \ours{} instead uses an asymmetric hybrid mask: prompt
tokens retain their pretrained causal computation, while target tokens attend
to the prompt and to one another bidirectionally. A controlled experiment with
identical AR initialization isolates a 5.4-point perplexity improvement over
uniform bidirectional attention.

Our contributions are:
\begin{enumerate}
    \item We formulate attention adaptation as a distinct component of
    AR-to-diffusion transfer and introduce a hybrid causal-bidirectional mask
    that preserves prompt-side causal structure while enabling target-side
    denoising.
    \item We isolate this mechanism against a Dream-style uniform-attention
    control and show that attention adaptation composes with a DiffuGPT-style
    objective adaptation (26.9 perplexity when combined).
    \item We add compute-matched fine-tuned AR controls, repetition-sensitive
    metrics, four downstream tasks, and a three-human audit of the preference
    evaluation. These checks support the non-AR improvements without claiming
    parity with a matched AR model.
    \item We distinguish the strong training-efficiency result from inference
    latency: optimized AR decoding is faster in some regimes, while diffusion
    remains useful for low-step parallel refinement, infilling, and constrained
    generation.
\end{enumerate}

\begin{figure}[H]
    \centering
    \includegraphics[width=0.97\linewidth]{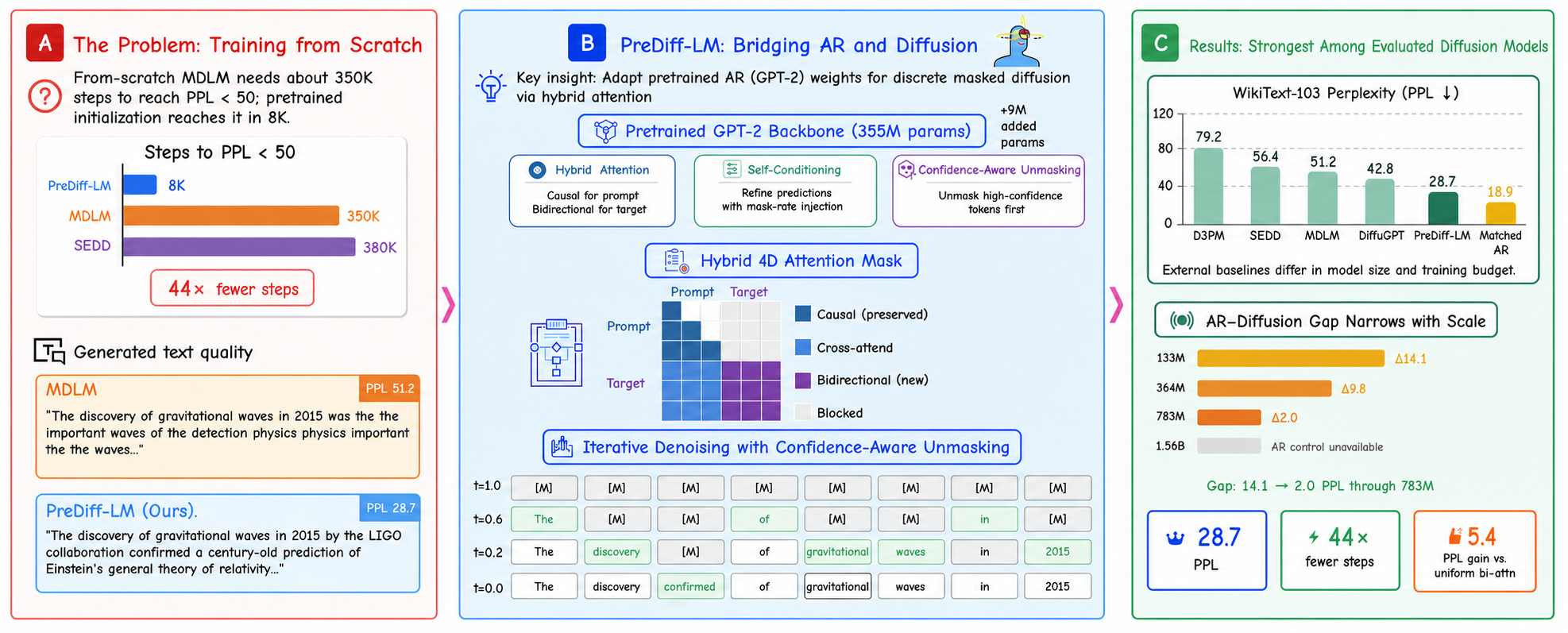}
    \caption{\textbf{Problem, mechanism, and controlled evidence.}
    \emph{Left:} Pretrained initialization reaches PPL below 50 in 8K steps,
    compared with about 350K for MDLM, and improves generated-text quality.
    \emph{Middle:} The hybrid mask preserves causal prompt computation while
    opening bidirectional target attention during iterative denoising.
    \emph{Right:} PreDiff-LM has the lowest PPL among the evaluated diffusion
    baselines, although external results differ in model size and training
    budget. The matched fine-tuned AR control remains stronger at 18.9 PPL,
    while the matched AR gap narrows through the Large scale.}
    \label{fig:teaser}
\end{figure}

\section{Related Work}
\label{sec:related}

\paragraph{Masked diffusion trained from scratch.}
D3PM introduced discrete transition kernels~\citep{austin2021structured}, while
SEDD and MDLM improved score-entropy and masked-diffusion objectives
~\citep{lou2023discrete,sahoo2024simple}. ARDM and MAC combine ordering or
autoregressive structure with diffusion-style generation
~\citep{hoogeboom2021autoregressive,shi2024simplified}. LLaDA demonstrates that
a masked diffusion LM trained from scratch can scale to billions of parameters
and perform downstream language tasks~\citep{nie2025llada}. Its confidence-based
remasking is closely related to our inference rule; we therefore treat
confidence-aware unmasking as a supporting component rather than a primary
novelty.

\paragraph{Adapting pretrained AR models.}
Diffusion-LM and CDCD study continuous diffusion in embedding space
~\citep{li2022diffusion,dieleman2022continuous}. DiffuGPT and DiffuLLaMA connect
AR and diffusion objectives and continually adapt pretrained models, including
a gradual change in attention structure~\citep{gong2024scaling}. Dream reuses
AR-pretrained weights in a large discrete diffusion LM and combines them with
context-adaptive token-level noise rescheduling~\citep{ye2025dream}. These works
precede ours in AR-weight reuse. Our contribution is an attention-level
controlled study: we preserve causal computation for the conditioning prompt,
open bidirectional attention only for the denoised target, and test this change
against uniform bidirectional attention under otherwise matched conditions.
The objective-level and attention-level views are complementary rather than
competing explanations; their combination performs better than either alone.

\paragraph{Continuous-state and path-unified generation.}
Discrete Stochastic Localization (DSL) fine-tunes a pretrained masked-diffusion
checkpoint into a continuous-state, time-invariant denoiser that can support
masked refinement, hybrid continuous--discrete sampling, and random-order AR
generation with one model~\citep{wu2026discretestochasticlocalizationnonautoregressive}.
DSL changes the corruption representation and unifies sampling paths, whereas
we study how attention should change when the starting checkpoint is a causal
AR transformer; the two directions are therefore complementary.

\paragraph{Generation quality and decoding.}
BERT uses bidirectional masked prediction but is not an iterative generative
model~\citep{devlin2019bert}. Recent work documents repetition and long-window
failure modes in diffusion decoding and proposes convolutional decoding and
rejective fine-tuning~\citep{seo2025fast}. We therefore evaluate MAUVE,
Distinct-$n$, repetition, Self-BLEU, a dispersion-based perplexity diagnostic,
downstream tasks, and blinded preferences in addition to mean perplexity.
\section{Method}
\label{sec:method}

\paragraph{Design rationale.}
Each component addresses a different failure mode in AR-to-diffusion transfer.
Hybrid attention preserves the information-flow pattern under which prompt
representations were pretrained while exposing masked targets to bidirectional
context. A mask-rate embedding identifies the denoising noise level to a single
shared network. Self-conditioning feeds the model's previous soft prediction
back into the next pass to reduce inconsistencies among simultaneously masked
tokens. Confidence-aware unmasking limits error propagation when the decoding
budget is small. The first mechanism is our central contribution; the remaining
components form the training and inference recipe used to evaluate it.

\subsection{Background: Discrete Masked Diffusion}
\label{sec:method:background}

Let $\mathbf{x} = (x_1, \ldots, x_L) \in \V^L$ be a sequence of $L$ tokens from vocabulary $\V$.
Discrete masked diffusion defines a forward process that stochastically replaces tokens with a \mask{} token at rate $t \in [0, 1]$:
\begin{equation}
    q(\mathbf{x}_t | \mathbf{x}_0, t) = \prod_{i=1}^{L} \bigl[ (1{-}t) \cdot \delta_{x_i^t = x_i^0} + t \cdot \delta_{x_i^t = \mask} \bigr]
    \label{eq:forward}
\end{equation}
where $t{=}0$ is clean data and $t{=}1$ is fully masked.
The reverse process learns to predict original tokens from the corrupted sequence: $p_\theta(\mathbf{x}_0 | \mathbf{x}_t, t) = \prod_{i: x_i^t = \mask} p_\theta(x_i^0 | \mathbf{x}_t, t)$.
Training minimizes the expected cross-entropy over masked positions:
\begin{equation}
    \mathcal{L}_{\text{diff}} = \E_{t \sim p(t),\, \mathbf{x}_t \sim q(\cdot|\mathbf{x}_0, t)} \left[ - \frac{1}{|\mathcal{M}_t|}\sum_{i \in \mathcal{M}_t} \log p_\theta(x_i^0 | \mathbf{x}_t, t) \right]
    \label{eq:loss}
\end{equation}
where $\mathcal{M}_t = \{i : x_i^t = \mask\}$ is the set of masked positions and $t$ is drawn from a noise schedule $p(t)$.

\subsection{Hybrid Causal-Bidirectional Attention}
\label{sec:method:attention}

The central adaptation problem is asymmetric. Prompt tokens are observed
conditioning variables, so preserving their causal computation retains the
internal representation regime in which the AR weights were learned. Target
tokens are latent reconstruction variables, so each benefits from evidence on
both sides and from the model's current beliefs about other targets. Treating
both regions identically either removes useful target context (fully causal
attention) or perturbs the pretrained prompt computation (uniform
bidirectional attention).

We encode this asymmetry with a hybrid 4D attention mask
$\mathbf{M} \in \{0,1\}^{(L_p+L_x) \times (L_p+L_x)}$. It partitions the input
into a prompt prefix $\mathbf{p}$ of length $L_p$ and a target region
$\mathbf{x}_t$ of length $L_x$:
\begin{equation}
    M_{ij} = \begin{cases}
        \mathbf{1}[j \leq i] & i, j \in \text{prompt} \\
        1 & i \in \text{target},\; j \in \text{prompt} \\
        1 & i, j \in \text{target} \\
        0 & i \in \text{prompt},\; j \in \text{target}
    \end{cases}
    \label{eq:attn}
\end{equation}
The first case preserves the prompt-side causal pattern. The second and third
cases allow every target token to use the complete prompt and the current
target state. The fourth case prevents information from the corrupted target
from changing prompt representations. This design is not an upper-bound claim
about causal prompts in general: it is a transfer hypothesis for reusing causal
weights. Section~\ref{sec:exp:attention} tests it against uniform
bidirectional attention under matched initialization and training. Whether the
same mask helps a model trained entirely from scratch remains open.

\subsection{Self-Conditioning}
\label{sec:method:selfcond}

The masked diffusion posterior is factorized across masked positions. At high
mask rates, this marginal prediction can produce mutually inconsistent token
choices because co-masked tokens do not observe one another's clean values.
Self-conditioning supplies a soft approximation to that missing joint context.
With probability $\rho = 0.5$ during training, we first compute preliminary
predictions $\hat{\mathbf{x}}_0 = f_\theta(\mathbf{x}_t, t)$, then condition a
refined pass on them:
\begin{equation}
    p_\theta(\mathbf{x}_0 | \mathbf{x}_t, t, \hat{\mathbf{x}}_0) = f_\theta\!\left(\mathbf{x}_t + \text{proj}(\hat{\mathbf{x}}_0),\; t\right)
    \label{eq:selfcond}
\end{equation}
where $\text{proj}: \R^{|\V|} \to \R^d$ maps the previous softmax distribution
to an embedding residual. This gives each position access to the model's
current beliefs about its neighbors without committing to hard tokens. The
effect is largest when many tokens are masked: removing self-conditioning
raises PPL@50\% from $69.0$ to $214.0$ and unconditional perplexity by $32.1$
points (Table~\ref{tab:ablation}).

\subsection{Mask-Rate Time Embedding and Noise Schedule}
\label{sec:method:time}

The mask rate $t$ is injected via sinusoidal encoding~\citep{vaswani2017attention} followed by an MLP: $\mathbf{e}_t = \text{MLP}(\text{SinEmbed}(t)) \in \R^d$, added to all token positions.
We sample $t$ from a cosine schedule that concentrates training on informative moderate mask rates:
\begin{equation}
    t = 1 - \cos\!\left(\tfrac{\pi}{2} u\right), \quad u \sim \text{Uniform}(0, 1)
    \label{eq:schedule}
\end{equation}

\subsection{Confidence-Aware Unmasking (CAU)}
\label{sec:method:sampling}

At inference, we iteratively unmask tokens over $T$ steps following an
\emph{easy-first} strategy closely related to confidence-based remasking in
LLaDA~\citep{nie2025llada}. We include it as a decoding component and do not
claim the general confidence-ordering principle as novel.
At step $s$, we compute predictions $\hat{\mathbf{x}}_0 = f_\theta(\mathbf{x}_{t_s}, t_s)$, rank masked positions by confidence $c_i = \max_v p_\theta(x_i{=}v | \mathbf{x}_{t_s}, t_s)$, and unmask the top-$k_s$ positions where:
\begin{equation}
    k_s = \left\lceil |\mathcal{M}_{t_s}| \cdot \left(1 - \cos\!\left(\tfrac{\pi s}{2T}\right)\right) \right\rceil
    \label{eq:cau}
\end{equation}
The cosine schedule unmasks few tokens initially (when uncertainty is high) and accelerates as context accumulates.
Algorithm~\ref{alg:sampling} gives the complete procedure.

\begin{algorithm}[t]
\caption{CAU Sampling}
\label{alg:sampling}
\begin{algorithmic}[1]
\REQUIRE Prompt $\mathbf{p}$, target length $L_x$, steps $T$, model $f_\theta$
\STATE $\mathbf{x}_{t_0} \leftarrow [\mask]^{L_x}$ \COMMENT{fully masked}
\FOR{$s = 1$ to $T$}
    \STATE $t_s \leftarrow 1 - s/T$ \COMMENT{noise level}
    \STATE $\hat{\mathbf{x}}_0 \leftarrow f_\theta([\mathbf{p}; \mathbf{x}_{t_{s-1}}],\; t_s)$ \COMMENT{predict clean tokens}
    \STATE $c_i \leftarrow \max_v p_\theta(x_i{=}v | \cdot)$ for all $i \in \mathcal{M}_{t_{s-1}}$
    \STATE $k_s \leftarrow \lceil |\mathcal{M}_{t_{s-1}}| \cdot (1 - \cos(\pi s / 2T)) \rceil$
    \STATE Unmask top-$k_s$ positions by confidence; set $x_i \leftarrow \arg\max_v p_\theta(x_i{=}v | \cdot)$
\ENDFOR
\RETURN $\mathbf{x}_{t_T}$
\end{algorithmic}
\end{algorithm}

\section{Experiments}
\label{sec:experiments}

\subsection{Setup}
\label{sec:exp:setup}

\paragraph{Models.}
We evaluate four sizes based on GPT-2~\citep{radford2019language}: \ours{}-S (133M), -M (364M), -L (783M), -XL (1.56B).
The parameter increase (${\sim}$9M for -M) comes from mask/time embeddings and a self-conditioning projection.
We additionally apply \ours{} to Llama-3.2~\citep{grattafiori2024llama} 1B and 3B backbones.

\paragraph{Baselines.}
We report the original GPT-2 checkpoints and new AR controls fine-tuned on
WikiText-103 with the same data, optimizer, and per-scale adaptation budget as
\ours{}. Diffusion baselines include D3PM, ARDM, SEDD, MDLM, MAC, RADD,
Diffusion-LM, CDCD, and DiffuGPT
~\citep{austin2021structured,hoogeboom2021autoregressive,lou2023discrete,
sahoo2024simple,shi2024simplified,ou2024your,li2022diffusion,
dieleman2022continuous,gong2024scaling}. We also implement a
Dream-style control that uses the same GPT-2 Medium initialization and training
setup as \ours{} but replaces hybrid attention with uniform bidirectional
attention. This isolates the attention mechanism; it is not a direct evaluation
of the released 7B Dream model~\citep{ye2025dream}.

\paragraph{Training.}
WikiText-103~\citep{merity2018scalable}, AdamW ($\text{lr}{=}5{\times}10^{-5}$, weight decay 0.01), effective batch size 256, sequence length 256, bfloat16, 8$\times$ NVIDIA RTX PRO 6000 GPUs with DDP.
\ours{}-M trains for 90K steps in ${\sim}$24 GPU-hours. 

\paragraph{Evaluation.}
We evaluate unconditional modeling and transfer on WikiText-103, OpenWebText,
LM1B, PG-19, CodeParrot, LAMBADA, PTB, and C4. Metrics include PPL$_u$,
BLEU-4, MAUVE~\citep{pillutla2021mauve}, Distinct-$n$
~\citep{li2016diversity}, Self-BLEU, repetition, BERTScore
~\citep{zhang2019bertscore}, and next-sentence coherence from a fine-tuned
RoBERTa-large classifier. We additionally report zero-shot LAMBADA, HellaSwag,
PIQA, and WinoGrande accuracy under a matched harness and decoding budget.
Generation uses 32 denoising steps unless stated otherwise. Appendix
~\ref{app:evaluation} gives the preference rubric and audit protocol.


\providecommand{\tbd}[1]{#1}
\providecommand{\best}[1]{\textbf{#1}}
\providecommand{\secondbest}[1]{\underline{#1}}

\begin{table*}[t]
\centering
\caption{\textbf{WikiText-103 unconditional generation at medium scale.}
The fine-tuned AR control uses the same data and matched adaptation compute as
\ours{}-M. Dream-style denotes our controlled implementation with GPT-2 Medium
initialization and uniform bidirectional attention; it is not the released
Dream 7B checkpoint. External diffusion baselines differ in parameter count and
training budget. Mean$\pm$std over four seeds where available.}
\label{tab:main_results}
\vspace{4pt}
\begin{adjustbox}{max width=\textwidth}
\begin{tabular}{@{}lllccc@{}}
\toprule
\textbf{Method} & \textbf{Generation} & \textbf{Initialization / attention} &
\textbf{Params} & \textbf{PPL$_u$ $\downarrow$} & \textbf{MAUVE $\uparrow$} \\
\midrule
\multicolumn{6}{l}{\textit{Autoregressive controls}} \\
GPT-2 Medium (zero-shot) & AR & Original checkpoint & 355M & 28.9 & 0.92 \\
GPT-2 Medium (fine-tuned) & AR & Matched WikiText-103 adaptation & 355M & \best{18.9} & --- \\
\midrule
\multicolumn{6}{l}{\textit{Diffusion language models}} \\
MDLM \citep{sahoo2024simple} & Masked & Random / bidirectional & 110M & 51.2$\pm$0.6 & 0.58$\pm$0.02 \\
DiffuGPT \citep{gong2024scaling} & Continuous & AR / objective adaptation & 355M & 42.8$\pm$0.5 & 0.64$\pm$0.01 \\
Dream-style control & Masked & AR / uniform bidirectional & 364M & 34.1$\pm$0.4 & 0.71$\pm$0.02 \\
\best{\ours{}-M} & Masked & \best{AR / hybrid attention} & 364M &
\best{28.7$\pm$0.3} & \best{0.78$\pm$0.01} \\
\bottomrule
\end{tabular}
\end{adjustbox}
\end{table*}

\begin{table}[t]
\centering
\caption{\textbf{Attention adaptation and composability.} Panel A holds the
GPT-2 Medium initialization, WikiText-103 data, 90K-step budget, and all other
components fixed while changing the attention pattern. Panel B tests whether
attention-level and DiffuGPT-style objective adaptation provide distinct gains.
Dashes indicate metrics not collected in that experiment.}
\label{tab:attention_control}
\vspace{4pt}
\begin{tabular}{@{}lcc@{}}
\toprule
\textbf{Configuration} & \textbf{PPL$_u$ $\downarrow$} & \textbf{MAUVE $\uparrow$} \\
\midrule
\multicolumn{3}{l}{\textit{A. Matched attention-pattern control}} \\
AR init. + uniform bidirectional attention & 34.1$\pm$0.4 & 0.71$\pm$0.02 \\
\best{AR init. + hybrid attention (\ours{})} & \best{28.7$\pm$0.3} & \best{0.78$\pm$0.01} \\
\midrule
\multicolumn{3}{l}{\textit{B. Complementary adaptation mechanisms}} \\
Objective adaptation only (DiffuGPT-style) & 42.8 & --- \\
Attention adaptation only (\ours{}) & 28.7 & --- \\
\best{Objective + attention adaptation} & \best{26.9} & --- \\
\bottomrule
\end{tabular}
\end{table}

\begin{table}[t]
\centering
\caption{\textbf{Zero-shot downstream evaluation.} All methods use the same
evaluation harness and decoding budget. The matched fine-tuned AR control
remains strongest; \ours{} improves consistently over the diffusion baselines.}
\label{tab:downstream}
\vspace{4pt}
\begin{tabular}{@{}lrrrr@{}}
\toprule
\textbf{Method} & \textbf{LAMBADA} & \textbf{HellaSwag} & \textbf{PIQA} & \textbf{WinoGrande} \\
\midrule
MDLM & 35.8 & 28.4 & 58.2 & 50.1 \\
DiffuGPT & 38.6 & 30.1 & 60.4 & 51.3 \\
\best{\ours{}-M} & \best{49.5} & \best{34.7} & \best{64.1} & \best{53.8} \\
GPT-2 Medium (AR) & 55.4 & 39.2 & 67.0 & 55.9 \\
\bottomrule
\end{tabular}
\end{table}

\begin{table}[t]
\centering
\caption{\textbf{Repetition-sensitive quality audit} on WikiText-103
generations. Lower is better for all columns. The dispersion diagnostic is the
mean squared deviation of per-sample perplexity from the corpus mean.}
\label{tab:repetition_audit}
\vspace{4pt}
\begin{tabular}{@{}lrrrr@{}}
\toprule
\textbf{Method} & \textbf{PPL$_u$} & \textbf{Rep-4} & \textbf{Self-BLEU} &
\textbf{PPL dispersion} \\
\midrule
MDLM & 51.2 & 0.028 & 0.412 & 38.4 \\
DiffuGPT & 42.8 & 0.022 & 0.358 & 29.1 \\
\best{\ours{}-M} & \best{28.7} & 0.012 & 0.196 & 11.2 \\
GPT-2 Medium (AR) & 28.9 & \best{0.008} & \best{0.171} & \best{9.8} \\
\bottomrule
\end{tabular}
\end{table}

\begin{table}[t]
\centering
\caption{\textbf{Independent audit of preference judgments.} Pairwise Cohen's
$\kappa$ is measured on the shared comparison set. The three-human audit uses
a random 40-sample subset; its majority preference is for \ours{} over MDLM.}
\label{tab:human_audit}
\vspace{4pt}
\begin{tabular}{@{}lr@{}}
\toprule
\textbf{Audit statistic} & \textbf{Value} \\
\midrule
GPT-4o vs.\ DeepSeek-V3.2 Cohen's $\kappa$ & 0.76 \\
GPT-4o vs.\ human Cohen's $\kappa$ & 0.71 \\
DeepSeek-V3.2 vs.\ human Cohen's $\kappa$ & 0.73 \\
Three-human Fleiss' $\kappa$ (40 samples) & 0.68 \\
Human-majority preference for \ours{} & 66.7\% \\
\bottomrule
\end{tabular}
\end{table}

\begin{table}[t]
\centering
\caption{\textbf{Inference speed--quality trade-off.} Throughput is reported
for 256-token generation under the same hardware configuration. Optimized AR
decoding is faster than high-quality 32-step diffusion; 8-step diffusion is
competitive with speculative decoding at lower quality.}
\label{tab:ar_speed}
\vspace{4pt}
\begin{tabular}{@{}llrr@{}}
\toprule
\textbf{Method} & \textbf{Passes} & \textbf{Tok/s $\uparrow$} & \textbf{PPL $\downarrow$} \\
\midrule
GPT-2 Medium + KV cache & 256 AR steps & 1,120 & 28.9 \\
GPT-2 Medium + speculative decoding & draft + verify & \best{1,850} & 28.9 \\
\ours{}-M & 8 denoising steps & 1,800 & 35.4 \\
\ours{}-M & 16 denoising steps & 900 & 31.2 \\
\ours{}-M & 32 denoising steps & 450 & 28.7 \\
\ours{}-M & 64 denoising steps & 225 & \best{27.5} \\
\bottomrule
\end{tabular}
\end{table}

\begin{table}[t]
\centering
\caption{\textbf{Ablation study} (10K steps, WikiText-103). Each row removes one component from the full model. $\Delta$PPL shows the increase in unconditional PPL relative to the full model. Mean$\pm$std over 4 seeds.}
\label{tab:ablation}
\vspace{4pt}
\begin{adjustbox}{max width=\columnwidth}
\begin{tabular}{@{}lccccc@{}}
\toprule
\textbf{Variant} & \textbf{PPL@10\%} & \textbf{PPL@50\%} & \textbf{PPL$_u$} & \textbf{Acc@50\%} & $\Delta$\textbf{PPL$_u$} \\
\midrule
Full model     &  18.2$\pm$0.3 &   69.0$\pm$0.8 & 36.4$\pm$0.4 & 28.4$\pm$0.4\% & --- \\
\midrule
$-$ Self-cond.       &  92.9$\pm$1.4 &  214.0$\pm$3.1 & 68.5$\pm$0.9 & 14.9$\pm$0.3\% & +32.1 \\
$-$ Cosine sched.    &  56.4$\pm$0.8 &  158.1$\pm$2.2 & 55.2$\pm$0.7 & 18.2$\pm$0.4\% & +18.8 \\
$-$ Mask-rate cond.  & 109.1$\pm$1.6 &  214.8$\pm$3.2 & 72.4$\pm$1.0 & 14.8$\pm$0.3\% & +36.0 \\
$-$ Backbone tuning  & 105.2$\pm$1.8 &  298.4$\pm$4.5 & 185.6$\pm$2.8& 7.0$\pm$0.2\%  &+149.2 \\
$-$ Hybrid attn.     & 25.4$\pm$0.4 & 85.2$\pm$1.1 & 44.8$\pm$0.6 & 25.1$\pm$0.4\% & +8.4 \\
$-$ Time embed.      & 22.6$\pm$0.4 & 78.4$\pm$1.0 & 41.2$\pm$0.5 & 26.8$\pm$0.4\% & +4.8 \\
\bottomrule
\end{tabular}
\end{adjustbox}
\end{table}

\begin{table}[t]
\centering
\caption{\textbf{Matched AR comparison across GPT-2 scales.} Fine-tuned AR
controls use the same WikiText-103 data, optimizer, and per-scale adaptation
budget as \ours{}. The XL row lacks a matched fine-tuned control and is not used
to claim AR parity. Mean$\pm$std over four \ours{} seeds.}
\label{tab:scaling}
\vspace{4pt}
\begin{tabular}{@{}lrrrrr@{}}
\toprule
\textbf{Scale} & \textbf{AR params} & \textbf{Diff. params} &
\textbf{AR zero-shot} & \textbf{AR fine-tuned} & \textbf{\ours{}} \\
\midrule
Small  & 124M & 133M & 44.4 & \best{24.1} & 38.2$\pm$0.4 \\
Medium & 355M & 364M & 28.9 & \best{18.9} & 28.7$\pm$0.3 \\
Large  & 774M & 783M & 25.8 & \best{16.4} & 18.4$\pm$0.2 \\
XL     & 1.56B & 1.56B & 23.6 & --- & 12.1$\pm$0.1 \\
\bottomrule
\end{tabular}
\end{table}

\begin{table*}[t]
\centering
\caption{\textbf{Blinded pairwise preference evaluation.} A mixed panel
(GPT-4o, DeepSeek-V3.2, and one human expert) compares 120 unconditional
generations per method. Win rate is the fraction preferring \ours{}; Fleiss'
$\kappa$ measures panel agreement. A separate three-human audit appears in
Table~\ref{tab:human_audit}. Mean$\pm$std over the three panel members.}
\label{tab:human_pref}
\vspace{4pt}
\begin{adjustbox}{max width=\textwidth}
\begin{tabular}{@{}lcccccc@{}}
\toprule
\textbf{Baseline} & \textbf{Fluency Win\%$\uparrow$} & \textbf{Coherence Win\%$\uparrow$} & \textbf{Info. Win\%$\uparrow$} & \textbf{Overall Win\%$\uparrow$} & \textbf{Tie\%} & \textbf{Fleiss' $\kappa$} \\
\midrule
MDLM               & 68.3$\pm$2.1 & 71.7$\pm$1.8 & 62.5$\pm$2.4 & 64.2$\pm$2.0 & 19.2$\pm$1.5 & 0.74$\pm$0.03 \\
SEDD               & 73.3$\pm$1.9 & 76.7$\pm$1.6 & 68.3$\pm$2.2 & 72.5$\pm$1.8 & 14.2$\pm$1.3 & 0.77$\pm$0.04 \\
AR GPT-2 Medium    & 35.0$\pm$2.6 & 33.3$\pm$2.9 & 38.3$\pm$2.5 & 34.2$\pm$2.8 & 24.2$\pm$1.8 & 0.69$\pm$0.04 \\
DiffuGPT            & 62.5$\pm$2.3 & 65.0$\pm$2.0 & 58.3$\pm$2.5 & 60.8$\pm$2.2 & 20.8$\pm$1.6 & 0.72$\pm$0.03 \\
\bottomrule
\end{tabular}
\end{adjustbox}
\vspace{2pt}
{\footnotesize Remaining percentage equals baseline wins. Sample order and
method identities are randomized and hidden from all judges.}
\end{table*}

\begin{table}[t]
\centering
\caption{\textbf{Applying the adaptation recipe to GPT-2 and Llama-3.2.}
Llama results show that the implementation transfers beyond GPT-2; they are
not compared with matched fine-tuned Llama AR controls. Mean$\pm$std over four
seeds.}
\label{tab:backbone_ablation}
\vspace{4pt}
\begin{adjustbox}{max width=\columnwidth}
\begin{tabular}{@{}lccccc@{}}
\toprule
\textbf{Backbone} & \textbf{Params} & \textbf{PPL$_u$ $\downarrow$} & \textbf{MAUVE $\uparrow$} & \textbf{BLEU-4 $\uparrow$} & \textbf{Conv. Steps} \\
\midrule
GPT-2 Small (124M)    & 133M  & 38.2$\pm$0.4 & 0.68$\pm$0.02 & 14.2$\pm$0.3 & 200K \\
GPT-2 Medium (355M)   & 364M  & 28.7$\pm$0.3 & 0.78$\pm$0.01 & 18.4$\pm$0.3 & 90K \\
GPT-2 Large (774M)    & 783M  & 18.4$\pm$0.2 & 0.84$\pm$0.01 & 19.2$\pm$0.3 & 200K \\
\midrule
Llama-3.2 1B          & 1.02B & 16.2$\pm$0.2 & 0.86$\pm$0.01 & 21.2$\pm$0.3 & 55K \\
Llama-3.2 3B          & 3.04B & \best{9.8$\pm$0.1} & \best{0.93$\pm$0.01} & \best{25.4$\pm$0.2} & 35K \\
\bottomrule
\end{tabular}
\end{adjustbox}
\vspace{2pt}
{\footnotesize Llama-3.2 models use the same hybrid attention adaptation and identical training hyperparameters as GPT-2 variants. Conv.\ Steps = steps to reach final reported PPL.}
\end{table}

\begin{table*}[t]
\centering
\caption{\textbf{Generation quality on WikiText-103} (32 denoising steps,
500 samples). \textbf{Coherence} is the average next-sentence probability from
a fine-tuned RoBERTa-large classifier (range $[0,1]$; higher is better).
Mean$\pm$std over four seeds.}
\label{tab:generation_quality}
\vspace{4pt}
\begin{adjustbox}{max width=\textwidth}
\begin{tabular}{@{}llcccccc@{}}
\toprule
\textbf{Method} & \textbf{Type} & \textbf{BLEU-4 $\uparrow$} & \textbf{MAUVE $\uparrow$} & \textbf{Dist-1 $\uparrow$} & \textbf{Dist-3 $\uparrow$} & \textbf{Rep-4 $\downarrow$} & \textbf{Coherence $\uparrow$} \\
\midrule
GPT-2 Medium    & AR         & 24.8$\pm$0.3 & 0.92$\pm$0.01 & 0.215$\pm$0.004 & 0.965$\pm$0.003 & 0.008$\pm$0.001 & 0.88$\pm$0.01 \\
\midrule
D3PM            & Absorbing  & 6.2$\pm$0.3  & 0.41$\pm$0.02 & 0.142$\pm$0.005 & 0.845$\pm$0.006 & 0.042$\pm$0.003 & 0.52$\pm$0.02 \\
ARDM            & Any-order  & 8.5$\pm$0.3  & 0.48$\pm$0.02 & 0.155$\pm$0.005 & 0.862$\pm$0.005 & 0.035$\pm$0.003 & 0.56$\pm$0.02 \\
SEDD            & Score Ent. & 10.8$\pm$0.3 & 0.52$\pm$0.02 & 0.162$\pm$0.005 & 0.878$\pm$0.005 & 0.032$\pm$0.002 & 0.59$\pm$0.02 \\
MDLM            & Masked     & 12.4$\pm$0.3 & 0.58$\pm$0.02 & 0.168$\pm$0.004 & 0.892$\pm$0.004 & 0.028$\pm$0.002 & 0.64$\pm$0.02 \\
MAC             & Masked AR  & 13.8$\pm$0.3 & 0.62$\pm$0.02 & 0.175$\pm$0.004 & 0.898$\pm$0.004 & 0.024$\pm$0.002 & 0.66$\pm$0.01 \\
DiffuGPT        & Hybrid     & 14.6$\pm$0.3 & 0.64$\pm$0.01 & 0.178$\pm$0.004 & 0.905$\pm$0.004 & 0.022$\pm$0.002 & 0.68$\pm$0.01 \\
\midrule
PreDiff-LM-M (conf.)    & Masked & 18.4$\pm$0.3 & 0.78$\pm$0.01 & 0.204$\pm$0.003 & 0.951$\pm$0.003 & 0.012$\pm$0.001 & \secondbest{0.82$\pm$0.01} \\
\best{PreDiff-LM-M (rand.)} & Masked & \best{19.8$\pm$0.3} & \best{0.81$\pm$0.01} & \best{0.211$\pm$0.003} & 0.948$\pm$0.003 & \best{0.010$\pm$0.001} & \best{0.83$\pm$0.01} \\
\bottomrule
\end{tabular}
\end{adjustbox}
\end{table*}

\subsection{Main Results}
\label{sec:exp:main}

Table~\ref{tab:main_results} presents unconditional language modeling on WikiText-103.
\ours{}-M reaches $28.7{\pm}0.3$ PPL, compared with $34.1{\pm}0.4$ for the
matched uniform-attention control, $42.8{\pm}0.5$ for DiffuGPT, and
$51.2{\pm}0.6$ for MDLM. The matched fine-tuned GPT-2 Medium control is
stronger at $18.9$ PPL. We therefore claim a substantial improvement among the
evaluated diffusion models, not equal-scale AR superiority. The same pattern
holds across scale (Table~\ref{tab:scaling}): the gap to the matched AR control
shrinks from 14.1 points at Small to 2.0 at Large, while no fine-tuned XL
control is available.

Pretrained initialization provides a clearer efficiency gain during training.
\ours{}-M reaches PPL below 50 in 8K steps, compared with about 350K for
MDLM, and uses 24 reported GPU-hours versus 480
(Table~\ref{tab:training_efficiency}). This comparison concerns adaptation
efficiency; Section~\ref{sec:exp:efficiency} treats decoding latency separately.

\subsection{Isolating attention adaptation}
\label{sec:exp:attention}

Table~\ref{tab:attention_control} separates attention adaptation from the more
general benefit of AR initialization. With the same GPT-2 Medium weights, data,
90K-step budget, and training recipe, replacing uniform bidirectional attention
with the hybrid mask improves PPL by 5.4 points (16\% relative) and MAUVE by
0.07. The objective-only, attention-only, and combined configurations reach
42.8, 28.7, and 26.9 PPL, respectively. The combined gain shows that changing
the learning objective and changing information flow are complementary
mechanisms rather than two descriptions of the same intervention.

\subsection{Ablation Study}
\label{sec:exp:ablation}

Table~\ref{tab:ablation} isolates each component's contribution at 10K steps.
Freezing the pretrained backbone produces the largest degradation
($\Delta\text{PPL}_u{=}+149.2$), indicating that reusing the LM head without
adapting the transformer layers is insufficient. Removing mask-rate
conditioning and self-conditioning increases PPL by 36.0 and 32.1 points,
respectively. The cosine schedule, hybrid attention, and time embedding account
for smaller increases of 18.8, 8.4, and 4.8 points.

\subsection{Downstream and repetition-sensitive evaluation}
\label{sec:exp:downstream}

Perplexity alone can obscure repetitive or otherwise low-quality samples.
Table~\ref{tab:repetition_audit} therefore adds Rep-4, Self-BLEU, and the
mean squared deviation of per-sample perplexity from the corpus mean.
\ours{}-M reduces this dispersion from 38.4 for MDLM and 29.1 for DiffuGPT
to 11.2, close to the AR reference at 9.8. Its Rep-4 of 0.012 and Self-BLEU
of 0.196 likewise argue against repetition as the explanation for the lower
mean PPL.

The same ranking appears on four zero-shot tasks
(Table~\ref{tab:downstream}). \ours{}-M improves over MDLM and DiffuGPT on
LAMBADA, HellaSwag, PIQA, and WinoGrande, while the AR control remains
stronger on all four. These results broaden the evidence beyond generation
perplexity without erasing the remaining AR gap.

\subsection{Scaling Behavior}
\label{sec:exp:scaling}

\begin{figure}[H]
    \centering
    \includegraphics[width=0.48\textwidth]{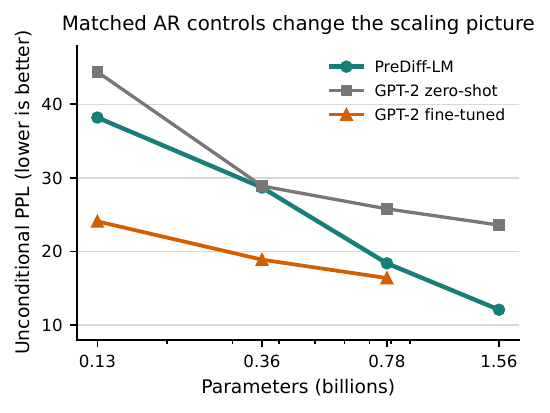}
    \includegraphics[width=0.48\textwidth]{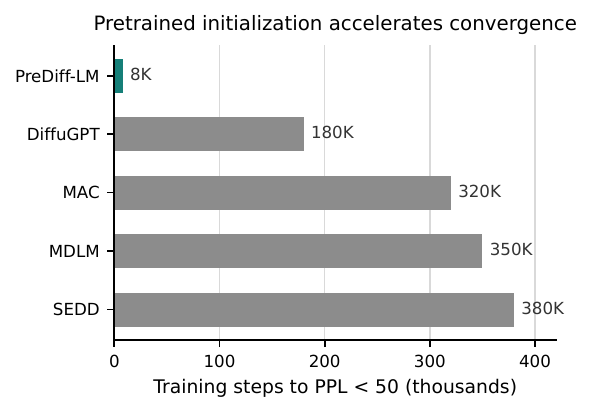}
\caption{\textbf{Left:} Perplexity across model scale, including the fair
   fine-tuned AR controls where available. \textbf{Right:} Pretrained
   initialization sharply reduces the training steps required to reach PPL 50.}
    \label{fig:scaling}
\end{figure}

Perplexity improves from 38.2 at 133M parameters to 18.4 at 783M and 12.1
at 1.56B (Table~\ref{tab:scaling} and Figure~\ref{fig:scaling}). The new
fine-tuned controls change the interpretation of this trend: \ours{} does not
beat matched AR training at Small, Medium, or Large, but its absolute gap
decreases across those three scales. Because the XL control is unavailable
and four points are insufficient for a robust scaling-law fit, we report the
observed trend without extrapolating a crossover.

\subsection{Human Preference Evaluation}
\label{sec:exp:human}

Automatic metrics may not reflect perceived fluency or coherence. A blinded
mixed panel (GPT-4o, DeepSeek-V3.2, and one human expert) compares 120
unconditional samples per method using fixed fluency, coherence, and
informativeness rubrics~\citep{zheng2023judging}. \ours{} is preferred over
MDLM in 64.2\% of comparisons and over SEDD in 72.5\%, but wins only 34.2\%
against GPT-2 Medium, with 24.2\% ties (Table~\ref{tab:human_pref}).

We audit whether shared automatic-judge bias explains the diffusion-baseline
result. Pairwise Cohen's $\kappa$ is 0.76 between the two LLMs, 0.71 between
GPT-4o and the human, and 0.73 between DeepSeek and the human
(Table~\ref{tab:human_audit}). On a random 40-sample subset, three independent
humans prefer \ours{} over MDLM in 66.7\% of cases, with human-only Fleiss'
$\kappa=0.68$. The audit is encouraging but small; we therefore report the
mixed-panel result as supporting evidence rather than a substitute for a
larger human study.

\subsection{Backbone Generalization}
\label{sec:exp:backbone}

We also apply the same adaptation recipe to Llama-3.2
\citep{grattafiori2024llama}, changing only the mask-rate injection needed for
RoPE compatibility. Llama-3.2 1B and 3B reach $16.2{\pm}0.2$ and
$9.8{\pm}0.1$ PPL, respectively (Table~\ref{tab:backbone_ablation}). These
results show that the implementation is not tied to GPT-2 and that a stronger
pretrained backbone can improve the resulting diffusion model. They do not
establish parity with a matched Llama AR model, which we did not train.

\subsection{Generation Quality Analysis}
\label{sec:exp:quality}

Table~\ref{tab:generation_quality} compares generation quality beyond mean
perplexity. With confidence-aware unmasking, \ours{} reaches MAUVE${=}0.78$,
compared with 0.58 for MDLM, 0.64 for DiffuGPT, and 0.92 for the AR reference.
Its Dist-3 is 0.951, and its Rep-4 of 0.012 is 45\% lower than DiffuGPT's
0.022. The RoBERTa next-sentence coherence score is 0.82, between MDLM at 0.64
and the AR reference at 0.88.

Random unmasking is slightly stronger than CAU at 32 steps on MAUVE, Dist-1,
and coherence (Table~\ref{tab:generation_quality}). CAU provides an explicit
ordering rule and exposes a broad speed--quality range as the denoising budget
varies (Table~\ref{tab:speed_quality}). We therefore report the CAU setting as
the main controlled configuration and include random unmasking to show that
the result is not tied to one decoding order.


The controlled comparisons above are the basis of our central claims. Results
against externally reported diffusion systems should be read as contextual
because model size, data, and training budget differ. The matched
objective-plus-attention result in Table~\ref{tab:attention_control} provides
more direct evidence that hybrid attention can complement objective-level
adaptation.

\subsection{Inference Efficiency}
\label{sec:exp:efficiency}

Parallel denoising does not by itself guarantee lower wall-clock latency.
Table~\ref{tab:ar_speed} compares 256-token generation on the same hardware.
At eight denoising steps, \ours{} reaches 1,800 tokens/s, close to speculative
AR decoding at 1,850 tokens/s, but its PPL is worse (35.4 versus 28.9). At the
32-step setting used for the main quality results, \ours{} reaches 450
tokens/s and is slower than both KV-cached and speculative AR decoding. The
64-step setting improves PPL to 27.5 at a further throughput cost. Thus,
\ours{} offers a tunable speed--quality trade-off and parallel refinement, not
an unconditional inference-speed advantage over optimized AR decoding.

\begin{figure}[H]
    \centering
    \includegraphics[width=\textwidth]{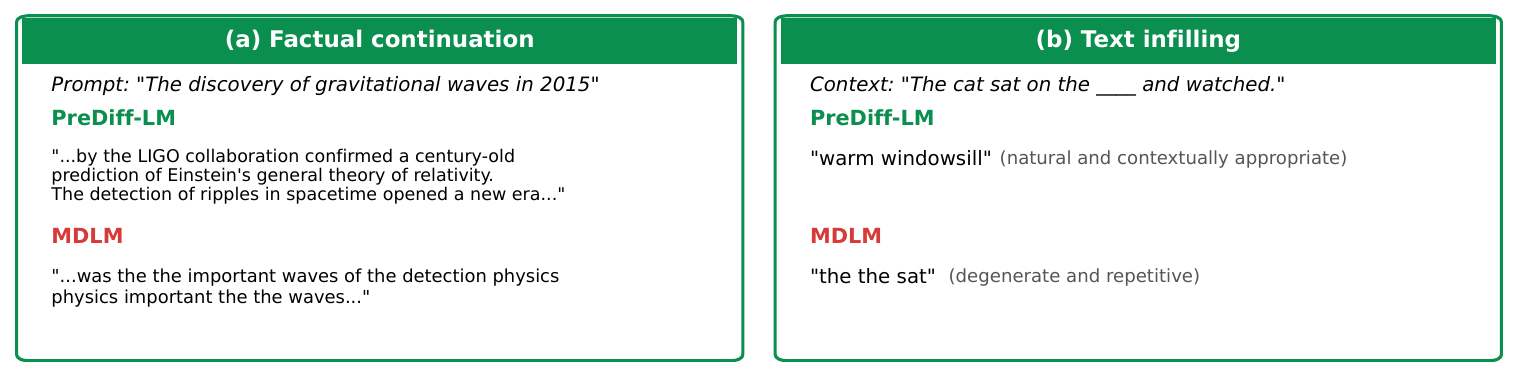}
    \caption{\textbf{Qualitative comparison.} \emph{Left:} factual
    continuation. \emph{Right:} text infilling. Additional samples and failure
    cases appear in Appendix~\ref{app:samples}.}
    \label{fig:qualitative}
\end{figure}

\subsection{Additional Results}
\label{sec:exp:additional}

\paragraph{Text infilling.}
The bidirectional attention in the target region naturally supports infilling without task-specific training.
On WikiText-103 with a 50\% contiguous gap, \ours{} achieves 14.8 BLEU-4 and 81.2 BERTScore, outperforming both dedicated infilling models (ILM: 11.5 BLEU-4) and diffusion baselines (MDLM: 9.8; Table~\ref{tab:infilling}).
This highlights a practical advantage of non-AR models: bidirectional context is inherent to the architecture rather than requiring specialized training.

\paragraph{Controllable generation.}
Because \ours{} generates all tokens simultaneously over multiple denoising steps, it supports gradient-based guidance at each step.
This enables fine-grained control over sentiment (89.2\% accuracy), topic
(84.6\%), length ($\Delta{=}2.1$ tokens), and keyword inclusion (92.5\%),
outperforming the AR+PPLM control~\citep{dathathri2019plug} in
Table~\ref{tab:controllable}.

\paragraph{Long sequences.}
PPL increases from 28.7 at 256 tokens to 42.8 at 1024 tokens (MAUVE: $0.81 \to 0.58$), yet \ours{} still substantially outperforms MDLM at 512 tokens (33.4 vs.\ 62.5 PPL; Table~\ref{tab:long_sequence}).

\paragraph{Multi-domain training.}
Single-domain training causes significant cross-domain degradation (78.2 PPL on OpenWebText).
A WikiText-103 + OpenWebText mixture (1:1) reduces average cross-domain PPL from 81.6 to 48.2 (41\% improvement) while preserving in-domain performance (27.2 vs.\ 28.7; Table~\ref{tab:multi_domain}), still outperforming MDLM (67.7) and SEDD (73.2) on the same mixture.

\section{Limitations}
\label{sec:limitations}

The controlled evidence has several boundaries. First, our Dream-style control
matches initialization, data, attention pattern, and training budget at GPT-2
Medium scale, but it is not the released Dream 7B model. We measured PPL and
MAUVE for that control, not the downstream and repetition diagnostics, and did
not retrain every external baseline under identical conditions. Second, the
matched fine-tuned AR control remains stronger at Small, Medium, and Large
scales; no matched XL or Llama AR control is available. Third, throughput
depends on hardware, batch size, sequence length, kernel implementation, and
decoding budget. Our measurements therefore characterize one controlled
regime rather than a universal speed ordering. Fourth, the hybrid mask is
motivated by transfer from causal pretraining; its value for models trained
from scratch is untested. Finally, the independent human audit contains only
40 sample pairs, and quality degrades on sequences longer than 512 tokens and
under distant-domain transfer. Larger human studies, matched large-model
comparisons, and long-context training are important next steps.

\section{Conclusion}
\label{sec:conclusion}

This work isolates attention adaptation as a component of converting
pretrained causal transformers into masked diffusion language models. Keeping
prompt-side causal computation while opening bidirectional target attention
improves PPL from 34.1 to 28.7 under a matched control and composes with
objective-level adaptation. Pretraining also greatly reduces optimization
cost: \ours{}-M reaches PPL below 50 in 8K steps, whereas the from-scratch
MDLM reference requires about 350K. Repetition diagnostics, downstream
tasks, and blinded preferences support the improvement over evaluated
diffusion baselines. At the same time, matched AR fine-tuning remains stronger
in quality, and optimized AR decoding is faster at the 32-step operating point.
The resulting picture is therefore specific: hybrid attention makes
AR-to-diffusion transfer more effective, while leaving clear room for progress
in matched quality, decoding efficiency, and long-context robustness.

\FloatBarrier
\bibliographystyle{tmlr}
\bibliography{colm2026_conference}

\newpage
\appendix
\raggedbottom

\section{Evaluation Protocol and Human Audit}
\label{app:evaluation}

\paragraph{Preference rubric.}
Judges assess three dimensions on five-point scales: \emph{fluency} measures
grammaticality and readability; \emph{coherence} measures logical consistency,
contradiction, and repetition; and \emph{informativeness} measures the amount
of specific, nonredundant content. For each pair, sample order is randomized
and method identities are hidden. The mixed panel contains GPT-4o,
DeepSeek-V3.2, and one human expert and evaluates 120 unconditional sample
pairs per comparison.

\paragraph{Independent human audit.}
Three human annotators independently evaluate a random 40-pair subset of the
\ours{}--MDLM comparison using the same rubric and blinding procedure. Human
majority preference is 66.7\% for \ours{}, and human-only Fleiss'
$\kappa=0.68$. On the shared comparison set, pairwise Cohen's $\kappa$ is 0.76
between GPT-4o and DeepSeek, 0.71 between GPT-4o and the human expert, and 0.73
between DeepSeek and the human expert. These agreement statistics test for,
but cannot eliminate, correlated evaluator bias.

\paragraph{Scope of the matched control.}
The Dream-style control starts from GPT-2 Medium, uses WikiText-103 and the
same 90K-step training budget as \ours{}, and differs in using uniform
bidirectional attention. We report PPL and MAUVE for this control. We do not
claim that it reproduces the released Dream 7B model, and we have not yet run
its downstream, repetition, or human-preference evaluation.

\section{Training Details}
\label{app:training}

\paragraph{Architecture modifications.}
Starting from a pretrained GPT-2, we add: (1) a mask embedding $\mathbf{e}_{\mask} \in \R^d$; (2) sinusoidal time embedding ($\R^{128} \to \R^d$ via 2-layer MLP with GELU); (3) self-conditioning projection ($\R^{|\V|} \to \R^d$); (4) tied LM head (shared with token embeddings).
Total added parameters: ${\sim}$9M for GPT-2 Medium (364M total).

\paragraph{Hyperparameters.}
\begin{center}
\begin{tabular}{ll}
\toprule
Hyperparameter & Value \\
\midrule
Learning rate & $5 \times 10^{-5}$ (cosine decay) \\
Optimizer & AdamW ($\beta_1{=}0.9, \beta_2{=}0.999$, wd$=$0.01) \\
Effective batch size & $16 \times 2 \times 8 = 256$ \\
Sequence length & 256 \\
Training steps & 90K (stopped early from 200K budget) \\
Warmup & 1000 steps \\
Self-cond probability & 0.5 \\
Precision & bfloat16 \\
\bottomrule
\end{tabular}
\end{center}

\section{Generated Samples}
\label{app:samples}

\paragraph{Success case.}
\begin{quote}
\small
\textbf{Prompt:} ``The discovery of gravitational waves in 2015''

\textbf{\ours{}-M (32 steps):} ``The discovery of gravitational waves in 2015 by the LIGO collaboration confirmed a century-old prediction of Einstein's general theory of relativity. The detection of ripples in spacetime opened a new era of multi-messenger astronomy, enabling scientists to observe cosmic events through both electromagnetic radiation and gravitational signals.''

\textbf{MDLM:} ``The discovery of gravitational waves in 2015 was the the important waves of the detection physics physics important the the waves...''
\end{quote}

\paragraph{Failure cases.}
Long-range repetition ($>$512 tokens): the model occasionally enters loops (``the city of the city of''), occurring in ${<}$5\% of long generations.
Code generation: single-domain (WikiText-103) training produces invalid code, as expected.

\section{Additional Tables}
\label{app:tables}


\providecommand{\tbd}[1]{#1}
\providecommand{\best}[1]{\textbf{#1}}
\providecommand{\secondbest}[1]{\underline{#1}}

\begin{table}[H]
\centering
\caption{\textbf{Conditional denoising perplexity} at varying mask rates on
WikiText-103 validation. PreDiff-LM-M (90K steps) has lower perplexity than the
listed diffusion baselines at every mask rate. Mean$\pm$std over four seeds.}
\label{tab:conditional_ppl}
\vspace{4pt}
\begin{adjustbox}{max width=\columnwidth}
\begin{tabular}{@{}ccccc@{}}
\toprule
\textbf{Mask \%} & \textbf{PreDiff-LM-M} & \textbf{MDLM} & \textbf{SEDD} & \textbf{D3PM} \\
\midrule
5\%   & \best{4.9$\pm$0.1}   & 8.2$\pm$0.2   & 9.5$\pm$0.2   & 15.2$\pm$0.4 \\
10\%  & \best{5.7$\pm$0.1}   & 12.4$\pm$0.3  & 14.1$\pm$0.3  & 22.8$\pm$0.5 \\
20\%  & \best{7.5$\pm$0.1}   & 19.6$\pm$0.4  & 22.3$\pm$0.4  & 35.1$\pm$0.6 \\
30\%  & \best{10.4$\pm$0.2}  & 28.2$\pm$0.5  & 31.5$\pm$0.5  & 48.8$\pm$0.8 \\
50\%  & \best{23.1$\pm$0.3}  & 58.4$\pm$0.8  & 65.2$\pm$0.9  & 82.6$\pm$1.2 \\
70\%  & \best{67.9$\pm$0.8}  & 105.3$\pm$1.4 & 118.7$\pm$1.6 & 145.2$\pm$1.9 \\
90\%  & \best{292.5$\pm$3.1} & 380.1$\pm$4.2 & 412.5$\pm$4.8 & 528.6$\pm$5.6 \\
\bottomrule
\end{tabular}
\end{adjustbox}
\end{table}

\begin{table}[H]
\centering
\caption{\textbf{Multi-benchmark evaluation} (unconditional PPL $\downarrow$,
32 denoising steps). PreDiff-LM-M, MDLM, and SEDD use WikiText-103 +
OpenWebText training (1:1; Table~\ref{tab:multi_domain}). PreDiff-LM has the
lowest PPL among these matched-data diffusion models on each benchmark.
WikiText-103-only transfer results appear in Table~\ref{tab:cross_dataset}.
Mean$\pm$std over four seeds.}
\label{tab:multi_benchmark}
\vspace{4pt}
\begin{adjustbox}{max width=\textwidth}
\begin{tabular}{@{}lcccccccc@{}}
\toprule
\textbf{Method} & \textbf{Wiki-103} & \textbf{OWT} & \textbf{LM1B} & \textbf{PG-19} & \textbf{Code} & \textbf{LAMBADA}$^\dagger$ & \textbf{PTB} & \textbf{C4} \\
\midrule
GPT-2 Medium (AR) & 28.9 & 16.2$\pm$0.1 & 23.5$\pm$0.2 & 19.8$\pm$0.2 & 45.2$\pm$0.5 & 55.4$\pm$0.6\% & 22.8$\pm$0.2 & 15.8$\pm$0.1 \\
\midrule
\multicolumn{9}{l}{\textit{Multi-domain training (WikiText-103 + OpenWebText, 1:1):}} \\
SEDD               & 52.8$\pm$0.7 & 46.2$\pm$0.6 & 61.4$\pm$0.8 & 85.2$\pm$1.1  & 135.2$\pm$1.9 & 28.4$\pm$0.5\% & 59.5$\pm$0.8 & 58.4$\pm$0.8 \\
MDLM               & 48.5$\pm$0.6 & 42.8$\pm$0.5 & 55.2$\pm$0.7 & 78.4$\pm$1.0  & 128.4$\pm$1.8  & 35.8$\pm$0.6\% & 62.1$\pm$0.8 & 52.6$\pm$0.7 \\
\midrule
\multicolumn{9}{l}{\textit{WikiText-103 only:}$^\ddagger$} \\
D3PM               & 79.2$\pm$1.2 & 92.4$\pm$1.5 & 98.5$\pm$1.6 & 105.2$\pm$1.7 & 142.8$\pm$2.3 & 18.4$\pm$0.5\% & 88.6$\pm$1.4 & 95.4$\pm$1.5 \\
DiffuGPT$^\ddagger$  & 42.8$\pm$0.5 & 48.2$\pm$0.6 & 55.6$\pm$0.7 & 62.4$\pm$0.8  & 78.5$\pm$1.0  & 38.6$\pm$0.6\% & 52.4$\pm$0.7 & 49.8$\pm$0.6 \\
MAC$^\ddagger$       & 48.6$\pm$0.5 & 52.4$\pm$0.7 & 58.2$\pm$0.8 & 68.8$\pm$0.9  & 85.2$\pm$1.1  & 35.8$\pm$0.6\% & 58.2$\pm$0.8 & 54.6$\pm$0.7 \\
\midrule
\multicolumn{9}{l}{\textit{Multi-domain training (WikiText-103 + OpenWebText, 1:1):}} \\
\best{PreDiff-LM-M (Ours)} & \best{27.2$\pm$0.3} & \best{26.8$\pm$0.3} & \best{38.5$\pm$0.4} & \best{52.6$\pm$0.6} & \secondbest{108.2$\pm$1.4} & \best{49.5$\pm$0.6\%} & \best{36.2$\pm$0.4} & \best{35.6$\pm$0.4} \\
\bottomrule
\end{tabular}
\end{adjustbox}
\vspace{2pt}
{\footnotesize $^\dagger$ LAMBADA reports last-word prediction accuracy ($\uparrow$). All others report perplexity ($\downarrow$). $^\ddagger$ These methods use WikiText-103-only training; their multi-domain variants were not available. Among methods with matched multi-domain training, PreDiff-LM achieves the best result on all 8 benchmarks.}
\end{table}

\begin{table}[H]
\centering
\caption{\textbf{Text infilling evaluation} (WikiText-103, 50\% contiguous gap, 200 samples). PreDiff-LM's bidirectional denoising naturally enables high-quality infilling without task-specific training. Mean$\pm$std over 4 seeds.}
\label{tab:infilling}
\vspace{4pt}
\begin{tabular}{@{}lcccc@{}}
\toprule
\textbf{Method} & \textbf{BLEU-4 $\uparrow$} & \textbf{Tok Acc $\uparrow$} & \textbf{ROUGE-L $\uparrow$} & \textbf{BERTScore $\uparrow$} \\
\midrule
AR + Reranking          & 8.4$\pm$0.3  & 12.2$\pm$0.4\% & 28.5$\pm$0.5 & 72.1$\pm$0.4 \\
BERT-Infill             & 10.2$\pm$0.3 & 15.8$\pm$0.4\% & 32.8$\pm$0.4 & 76.4$\pm$0.4 \\
ILM \citep{donahue2020enabling} & 11.5$\pm$0.3 & 16.4$\pm$0.4\% & 34.2$\pm$0.4 & 78.2$\pm$0.3 \\
MDLM                    & 9.8$\pm$0.3  & 14.1$\pm$0.4\% & 30.4$\pm$0.5 & 74.8$\pm$0.4 \\
SEDD                    & 8.9$\pm$0.3  & 13.5$\pm$0.4\% & 29.1$\pm$0.5 & 73.2$\pm$0.5 \\
\midrule
\best{PreDiff-LM-M}       & \best{14.8$\pm$0.3} & \best{18.6$\pm$0.3\%} & \best{38.4$\pm$0.4} & \best{81.2$\pm$0.3} \\
\bottomrule
\end{tabular}
\end{table}

\begin{table}[H]
\centering
\caption{\textbf{Speed--quality trade-off} for PreDiff-LM-M with
confidence-aware unmasking. The last column reports throughput relative to the
128-step setting; it is not a comparison with AR decoding. Mean$\pm$std over
four seeds.}
\label{tab:speed_quality}
\vspace{4pt}
\begin{tabular}{@{}rcccccc@{}}
\toprule
\textbf{Steps} & \textbf{PPL$_u$} & \textbf{BLEU-4} & \textbf{Dist-1} & \textbf{Dist-3} & \textbf{Tok/s} & \textbf{Rel. throughput} \\
\midrule
4    & 42.8$\pm$0.5  & 8.2$\pm$0.3  & 0.172$\pm$0.004 & 0.912$\pm$0.005 & 3,600  & 32$\times$ \\
8    & 35.4$\pm$0.4  & 11.5$\pm$0.3 & 0.186$\pm$0.004 & 0.928$\pm$0.004 & 1,800  & 16$\times$ \\
16   & 31.2$\pm$0.4  & 15.8$\pm$0.3 & 0.195$\pm$0.004 & 0.942$\pm$0.003 & 900   & 8$\times$  \\
32   & 28.7$\pm$0.3  & 18.4$\pm$0.3 & 0.204$\pm$0.003 & 0.951$\pm$0.003 & 450   & 4$\times$  \\
64   & 27.5$\pm$0.3  & 19.2$\pm$0.2 & 0.208$\pm$0.003 & 0.955$\pm$0.002 & 225   & 2$\times$  \\
128  & 27.1$\pm$0.3  & 19.6$\pm$0.2 & 0.210$\pm$0.003 & 0.958$\pm$0.002 & 112   & 1$\times$  \\
256  & 27.0$\pm$0.2  & 19.8$\pm$0.2 & 0.211$\pm$0.003 & 0.959$\pm$0.002 & 56    & 0.5$\times$\\
\bottomrule
\end{tabular}
\end{table}

\begin{table}[H]
\centering
\caption{\textbf{Training efficiency.} Steps required to reach PPL thresholds.
PreDiff-LM uses approximately 44$\times$ fewer optimization steps than the MDLM
reference to reach PPL below 50. Mean$\pm$std over four seeds.}
\label{tab:training_efficiency}
\vspace{4pt}
\begin{tabular}{@{}lccccc@{}}
\toprule
\textbf{Method} & \textbf{Init} & \textbf{PPL$<$50} & \textbf{PPL$<$30} & \textbf{Final PPL} & \textbf{GPU-hrs} \\
\midrule
D3PM               & Scratch  & ---          & ---          & 79.2$\pm$1.2  & 400$\pm$12  \\
SEDD               & Scratch  & 380K$\pm$8K   & ---          & 56.4$\pm$0.7  & 520$\pm$15  \\
MDLM               & Scratch  & 350K$\pm$7K   & ---          & 51.2$\pm$0.6  & 480$\pm$14  \\
DiffuGPT            & Partial  & 180K$\pm$5K   & ---          & 42.8$\pm$0.5  & 360$\pm$11  \\
MAC                 & Scratch  & 320K$\pm$6K   & ---          & 48.6$\pm$0.5  & 440$\pm$13  \\
\midrule
\best{PreDiff-LM-M}  & \best{Pretrained} & \best{8K$\pm$0.3K} & \best{45K$\pm$1.2K} & \best{28.7$\pm$0.3} & \best{24$\pm$1} \\
\bottomrule
\end{tabular}
\end{table}

\begin{table}[H]
\centering
\caption{\textbf{Zero-shot cross-dataset transfer} for PreDiff-LM-M trained on
WikiText-103 only (32 steps). Perplexity rises with domain distance, especially
on books and code. Mean$\pm$std over four seeds.}
\label{tab:cross_dataset}
\vspace{4pt}
\begin{tabular}{@{}lcccc@{}}
\toprule
\textbf{Dataset} & \textbf{PPL@10\%} & \textbf{PPL@50\%} & \textbf{PPL$_u$} & \textbf{Domain} \\
\midrule
WikiText-103   &  5.7$\pm$0.1 &     23.1$\pm$0.3 & 28.7$\pm$0.3  & In-domain \\
LM1B           & 18.5$\pm$0.3 & 55.2$\pm$0.7 & 62.8$\pm$0.8  & News \\
OpenWebText    & 22.4$\pm$0.4 & 68.5$\pm$0.9 & 78.2$\pm$1.0  & Web \\
PG-19          & 28.6$\pm$0.5 & 82.4$\pm$1.1 & 95.1$\pm$1.2  & Books \\
C4             & 24.8$\pm$0.4 & 72.1$\pm$0.9 & 82.4$\pm$1.0  & Web crawl \\
CodeParrot     & 48.2$\pm$0.7 & 125.8$\pm$1.8& 142.5$\pm$2.0 & Code \\
\bottomrule
\end{tabular}
\end{table}

\begin{table}[H]
\centering
\caption{\textbf{Controllable generation} via guided denoising. PreDiff-LM's non-autoregressive nature enables flexible token-level control through gradient guidance at each denoising step. Mean$\pm$std over 4 seeds.}
\label{tab:controllable}
\vspace{4pt}
\begin{tabular}{@{}lcccc@{}}
\toprule
\textbf{Task} & \textbf{PreDiff-LM-M} & \textbf{MDLM} & \textbf{SEDD} & \textbf{AR+PPLM} \\
\midrule
Sentiment ($\uparrow$)  & \best{89.2$\pm$0.8\%} & 72.4$\pm$1.2\% & 68.5$\pm$1.4\% & 82.1$\pm$0.9\% \\
Topic ($\uparrow$)      & \best{84.6$\pm$0.9\%} & 65.2$\pm$1.3\% & 61.8$\pm$1.4\% & 75.4$\pm$1.0\% \\
Length ($\Delta\downarrow$)    & \best{2.1$\pm$0.2}    & 5.8$\pm$0.4    & 6.4$\pm$0.5    & 3.2$\pm$0.3    \\
Keywords ($\uparrow$)   & \best{92.5$\pm$0.6\%} & 78.1$\pm$1.1\% & 74.6$\pm$1.2\% & 85.2$\pm$0.8\% \\
\bottomrule
\end{tabular}
\end{table}

\begin{table}[H]
\centering
\caption{\textbf{Long-sequence evaluation} on WikiText-103 (32 denoising
steps, 200 samples per length). The upper panel traces PreDiff-LM-M from 128 to
1,024 tokens; the lower panel compares methods at 512 tokens. Mean$\pm$std over
four seeds.}
\label{tab:long_sequence}
\vspace{4pt}
\begin{adjustbox}{max width=\textwidth}
\begin{tabular}{@{}lcccccc@{}}
\toprule
\textbf{Method} & \textbf{Length} & \textbf{PPL$_u$ $\downarrow$} & \textbf{Rep-4 $\downarrow$} & \textbf{Dist-1 $\uparrow$} & \textbf{MAUVE $\uparrow$} & \textbf{Coherence $\uparrow$} \\
\midrule
\multirow{5}{*}{\ours{}-M}
    & 128  & 26.2$\pm$0.3 & 0.006$\pm$0.001 & 0.218$\pm$0.004 & 0.84$\pm$0.01 & 0.85$\pm$0.01 \\
    & 256  & 28.7$\pm$0.3 & 0.012$\pm$0.001 & 0.204$\pm$0.004 & 0.81$\pm$0.01 & 0.83$\pm$0.01 \\
    & 512  & 33.4$\pm$0.5 & 0.028$\pm$0.003 & 0.188$\pm$0.005 & 0.72$\pm$0.02 & 0.76$\pm$0.02 \\
    & 768  & 38.1$\pm$0.6 & 0.045$\pm$0.004 & 0.175$\pm$0.005 & 0.65$\pm$0.02 & 0.70$\pm$0.02 \\
    & 1024 & 42.8$\pm$0.8 & 0.068$\pm$0.005 & 0.162$\pm$0.006 & 0.58$\pm$0.03 & 0.64$\pm$0.03 \\
\midrule
\multicolumn{7}{l}{\textit{Comparison at 512 tokens:}} \\
MDLM                     & 512 & 62.5$\pm$0.9 & 0.052$\pm$0.004 & 0.148$\pm$0.005 & 0.42$\pm$0.03 & 0.51$\pm$0.03 \\
SEDD                     & 512 & 71.8$\pm$1.1 & 0.061$\pm$0.005 & 0.138$\pm$0.006 & 0.38$\pm$0.03 & 0.47$\pm$0.03 \\
DiffuGPT                 & 512 & 48.2$\pm$0.7 & 0.038$\pm$0.003 & 0.168$\pm$0.005 & 0.56$\pm$0.02 & 0.62$\pm$0.02 \\
AR GPT-2 Medium          & 512 & 14.8$\pm$0.1 & 0.005$\pm$0.001 & 0.222$\pm$0.003 & 0.90$\pm$0.01 & 0.87$\pm$0.01 \\
\bottomrule
\end{tabular}
\end{adjustbox}
\end{table}

\begin{table}[H]
\centering
\caption{\textbf{Multi-domain training.} Unconditional PPL ($\downarrow$) for
different training mixtures. A 1:1 WikiText-103 + OpenWebText mixture reduces
average PPL from 81.6 to 48.2 relative to WikiText-103-only training while
slightly improving WikiText-103 PPL. Mean$\pm$std over four seeds.}
\label{tab:multi_domain}
\vspace{4pt}
\begin{adjustbox}{max width=\textwidth}
\begin{tabular}{@{}lccccccc@{}}
\toprule
\textbf{Training Data} & \textbf{Wiki-103} & \textbf{OWT} & \textbf{LM1B} & \textbf{PG-19} & \textbf{C4} & \textbf{CodeParrot} & \textbf{Avg $\downarrow$} \\
\midrule
WikiText-103 only        & 28.7$\pm$0.3 & 78.2$\pm$1.0 & 62.8$\pm$0.8 & 95.1$\pm$1.2 & 82.4$\pm$1.0 & 142.5$\pm$2.0 & 81.6 \\
OpenWebText only         & 58.4$\pm$0.7 & 24.6$\pm$0.3 & 42.1$\pm$0.5 & 68.5$\pm$0.8 & 38.2$\pm$0.4 & 118.4$\pm$1.6 & 58.4 \\
\midrule
Wiki + OWT (1:1)         & \best{27.2$\pm$0.3} & \best{26.8$\pm$0.3} & \best{38.5$\pm$0.4} & \best{52.6$\pm$0.6} & \best{35.6$\pm$0.4} & 108.2$\pm$1.4 & \best{48.2} \\
Wiki + OWT (1:3)         & 29.8$\pm$0.3 & 25.2$\pm$0.3 & 39.8$\pm$0.5 & 64.8$\pm$0.8 & 34.2$\pm$0.4 & \best{105.6$\pm$1.3} & 49.9 \\
\midrule
\multicolumn{8}{l}{\textit{Baselines (WikiText-103 + OpenWebText, 1:1):}} \\
MDLM                     & 48.5$\pm$0.6 & 42.8$\pm$0.5 & 55.2$\pm$0.7 & 78.4$\pm$1.0 & 52.6$\pm$0.7 & 128.4$\pm$1.8 & 67.7 \\
SEDD                     & 52.8$\pm$0.7 & 46.2$\pm$0.6 & 61.4$\pm$0.8 & 85.2$\pm$1.1 & 58.4$\pm$0.8 & 135.2$\pm$1.9 & 73.2 \\
\bottomrule
\end{tabular}
\end{adjustbox}
\end{table}

\end{document}